\documentclass[9pt]{article}
\usepackage{amsmath,amssymb}
\usepackage{graphicx}
\usepackage{epstopdf}
\usepackage{qtree}
\usepackage{pict2e}
\usepackage{linguex, cgloss4e}
\usepackage{natbib}
\usepackage[latin1]{inputenc}

\usepackage{tree-dvips}
\begin{document}
\def\P{\mathbf{P}}
\def\Q{\mathbf{Q}}
\setlength{\parindent}{0cm}
\parskip 4mm
\pagestyle{empty}
\pagenumbering{arabic}
\pagestyle{headings}
\title{On the origin of ambiguity in efficient communication}
\author{Jordi Fortuny$^1$ and Bernat Corominas-Murtra$^{2,3}$  \\
\multicolumn{1}{p{.7\textwidth }}{\centering\emph{$^1$Departament de Filologia Catalana, Facultat de Filologia, Universitat de Barcelona, Gran Via de les Corts Catalanes, 585. Barcelona 08007, Spain}}\\
\multicolumn{1}{p{.7\textwidth }}{\centering\emph{$^2$ Section for Science of Complex Systems; Medical University of Vienna, Spitalgasse 23; A-1090, Austria\\
							$^3$ICREA-Complex Systems
 Lab,  Universitat Pompeu  Fabra,  Dr.  Aiguader  80, 08003  Barcelona,
  Spain	}}}

\maketitle
\newpage
\begin{abstract}
This article studies the emergence of ambiguity in communication through the concept of
logical  irreversibility  and   within  the  framework  of  Shannon's
information theory. This leads us to a precise and general expression of the intuition behind Zipf's vocabulary balance in terms of a symmetry equation between the complexities of the coding and the decoding processes
that imposes  an unavoidable amount of logical uncertainty in natural
communication. Accordingly, the emergence of irreversible computations
is  required  if the  complexities  of  the  coding and  the  decoding
processes are balanced  in a symmetric scenario, which  means that the
emergence of  ambiguous codes is  a necessary condition  for natural
communication to succeed. 
\end{abstract}
\tableofcontents
\newpage

\section{Introduction}
\label{intro}
It  is a  common  observation that  natural  languages are  ambiguous, namely, that  linguistic utterances can  potentially be assigned more than one  interpretation and that receivers of linguistic utterances need to  resort to  supplementary information  (i.e., the linguistic or the communicative   context)   to   choose   one   among   the  available interpretations. 

Both linguists and logicians have been interested in this observation. On the one hand, a traditional task  of grammar is to illustrate and classify ambiguity, which may be of different types; in this regard it is important to determine how  apparently ambiguous utterances are disambiguated at the  relevant level of  representation. Indeed, the search for a parsimonious treatment of certain ambiguities such as scope ambiguities has been one of the most powerful motors in the development of the formal inquiry of the syntax-semantics interface, since its modern inception in Montague's semiotic program \cite{montague1974}. It is no exaggeration at all, in our opinion, to say that the presence of ambiguity (particularly, scope ambiguity) and the apparent mismatch between the form and the alleged semantic structure of quantified expressions in natural languages have been the two major guiding problems in the development of a formal theory of the syntax-semantics interface of natural languages.  

On the other hand, logicians in general would not be as
interested   in  describing  or   characterizing  the   phenomenon  of
ambiguity as in the  construction of unambiguous artificial languages,
whose  primitive  symbols have  a  univocal  interpretation and  whose
formulae are constructed by  the appropriate recursive syntactic definitions
and unambiguously interpreted by the relevant compositional semantic rules, formulated as recursive definitions that trace back the syntactic construction of the formulae. Not surprisingly,  some   philosophers,  such as \cite[3.323-3.325]{wittgenstein}, identified  the ambiguity  of  ordinary  language   as  the  source  of  philosophical
confusion,  and  aspired  to  construct  a language  whose  signs  were
univocal  and whose  propositions  mirrored the  logical structure  of
reality itself. 

It is a rather common view that the presence of phenomena such as ambiguity and garden path sentences suggests that language is poorly designed for communication \cite{chomsky2008}. In fact, there are at least two opposite starting hypotheses about the nature and emergence of ambiguity in natural communication systems. It could well be that ambiguity is an intrinsic imperfection, since natural, self-organized codes of communication are not perfect, but coevolved through a fluctuating medium and no one designed them. But it may be as well that ambiguity is the result of an optimization process, inasmuch as natural, self-organized codes of communication {\em must} satisfy certain constraints that a non-ambiguous artificial language can afford to neglect. Logical languages, for instance, are constructed to study relations such as logical consequence and equivalence among well-formed formulae, for which it is \emph{desirable} to define syntactic rules that do not generate syntactically ambiguous expressions.\footnote{It is desirable, but not mandatory. As noted by Thomason in his `Introduction' \cite[Chapter 1]{montague1974}, ``[A] by-product of Montague's work (...) is a theory of how logical consequence can be defined for languages admitting syntactic ambiguity. For those logicians concerned only with artificial languages this generalization will be of little interest, since there is no serious point to constructing an artificial language that is not disambiguated (p.4, note 5)'' if the objective is to characterize logical notions such as consequence. However, this generalization is relevant for the development of `Universal Grammar' in Montague's sense, i.e., for the development of a general and uniform mathematical theory valid for the syntax and the semantics of both artificial and natural languages.} However, in the design of logical languages that provide the appropriate tools for that particular purpose, certain features that may be crucial in the emergence of natural communication systems are neglected, such as the importance of the cost in generating expressions and the role of cooperation between the coder and the decoder in the process of communicating those expressions \cite{wang2008}, a factor that is completely extraneous to the design of logical languages. 

Despite the indisputability of  ambiguity in  natural languages and  the attention
that this observation has  received among linguists, philosophers and
logicians, it is fair to conclude that the emergence of ambiguity has
not come yet under serious theoretical scrutiny.

In this article we provide a general mathematical definition of ambiguity through the computational concept of logical irreversibility and we quantify the ambiguity of a code in terms of the amount of uncertainty of the reversal of the coding process. 
 We finally capitalize on the two above-mentioned factors (the importance of the cost in generating expressions and the role of cooperation between the coder and the decoder) in order to provide a detailed argument for the idea that ambiguity is an unavoidable consequence of the following efficiency factor in natural communication systems: interacting communicative agents must attain a code that tends to minimize the complexities of both the coding and the decoding processes. 
As a proof of concept, we thoroughly explore a simple system based on two agents --coder and decoder-- under a symmetrical --cooperative-- scenario, and we show that ambiguity must emerge.

The remainder of this article is organized as follows. In section \ref{revers} we introduce Landauer's concept of logical (ir)reversibility of a given computational device and we quantify the degree of ambiguity of a code as the amount of uncertainty in reversing the coding process. In section \ref{LEP} we introduce Zipf's vocabulary balance condition, a particular instance of Zipf's Least Effort Principle \cite{zipf1965}, and we show how it can be properly generalized and accomodated to the information-theoretic framework adopted in section \ref{revers}. We conclude our reasoning by showing that, if the coding and the decoding
processes  are  performed  in  a  cooperative  regime  expressed in terms of a symmetry equation between coding and decoding complexities, a certain amount of logical uncertainty or ambiguity is unavoidable.  In section \ref{Discussion} we recapitulate our derivation of the presence of ambiguity and we stress a further important result intimately related to our development: Zipf's law, another well-known and ubiquitous feature of natural language products, is the sole expected outcome of an evolving communicative system that satisfies the symmetry equation between coding and decoding complexities, as argued in \cite{corominas2011}. Appendix \ref{Inequalities} emphasizes the relationship between logical irreversibility and thermodynamical irreversibility.

\section{Ambiguity and logical reversibility}
\label{revers}

In this section we begin by informally presenting the concept of {\em logical (ir)reversibility} of a given computation (subsection \ref{LI}). Subsequently we formally explain how a code generated through logically irreversible computations is necessarily ambiguous (subsection \ref{TM}) and we quantify the degree of ambiguity as the minimum amount of information needed to properly reconstruct a given message (subsection \ref{ambcodes}). 

\subsection{\textit{Logical irreversibility}}
\label{LI}

Traditionally, the concept of computation is theoretically studied as an abstract process of data manipulation and only its logical properties are taken into consideration; however, if we want to investigate how an abstract computation is realized by a physical system, such as an electronic machinery, an abacus or a biological system as the brain, it becomes important to consider the connections between the logical properties of (abstract) computations and the physical --or more precisely, thermodynamical-- properties of the system that performs those computations. 

The fundamental examination of the physical constraints that computations must satisfy when they are performed by a physical system was started by \cite{landauer1961}, and continued in several other works \cite{bennett1973}, \cite{bennettlandauer1985}, \cite{short2005}, \cite{bennett2008}, \cite{toffoli1980}. The general objective of these approaches is to determine the physical limits of the process of computing, the ``general laws that must govern all information processing no matter how it is accomplished'' \cite[p. 48]{bennettlandauer1985}. Accordingly, the concept of computation is subject to the same questions that apply to other physical processes, and thus the following questions become central to the physical study of computational devices \cite[p. 48]{bennettlandauer1985}: 

\begin{enumerate}
\item How much energy must be expended to perform a particular computation?
\label{energy}
\item How long must a computation take?
\label{length}
\item How large must the computing device be?
\label{size}
\end{enumerate}

It must be remarked that by answering these questions one expects to find fundamental physical principles that govern any kind of information processing, and not the actual limits of a particular techonological implementation. A central objective within this general framework is the study of the properties of reversibility and irreversibility relative to a computational process, a distinction that was first considered in relation to the problem of heat generation during the computational process. An important idea relative to question (\ref{energy}) is formulated in the so-called Landauer's principle, according to which any \emph{irreversible} computation generates an amount of heat (cfr. \cite{bennettlandauer1985} and also Appendix \ref{Inequalities}). Let us thus introduce the concept of logical reversibility/irreversibility, which is crucial to our concerns. 

\begin{figure}
	\centering
		\includegraphics[scale=0.4]{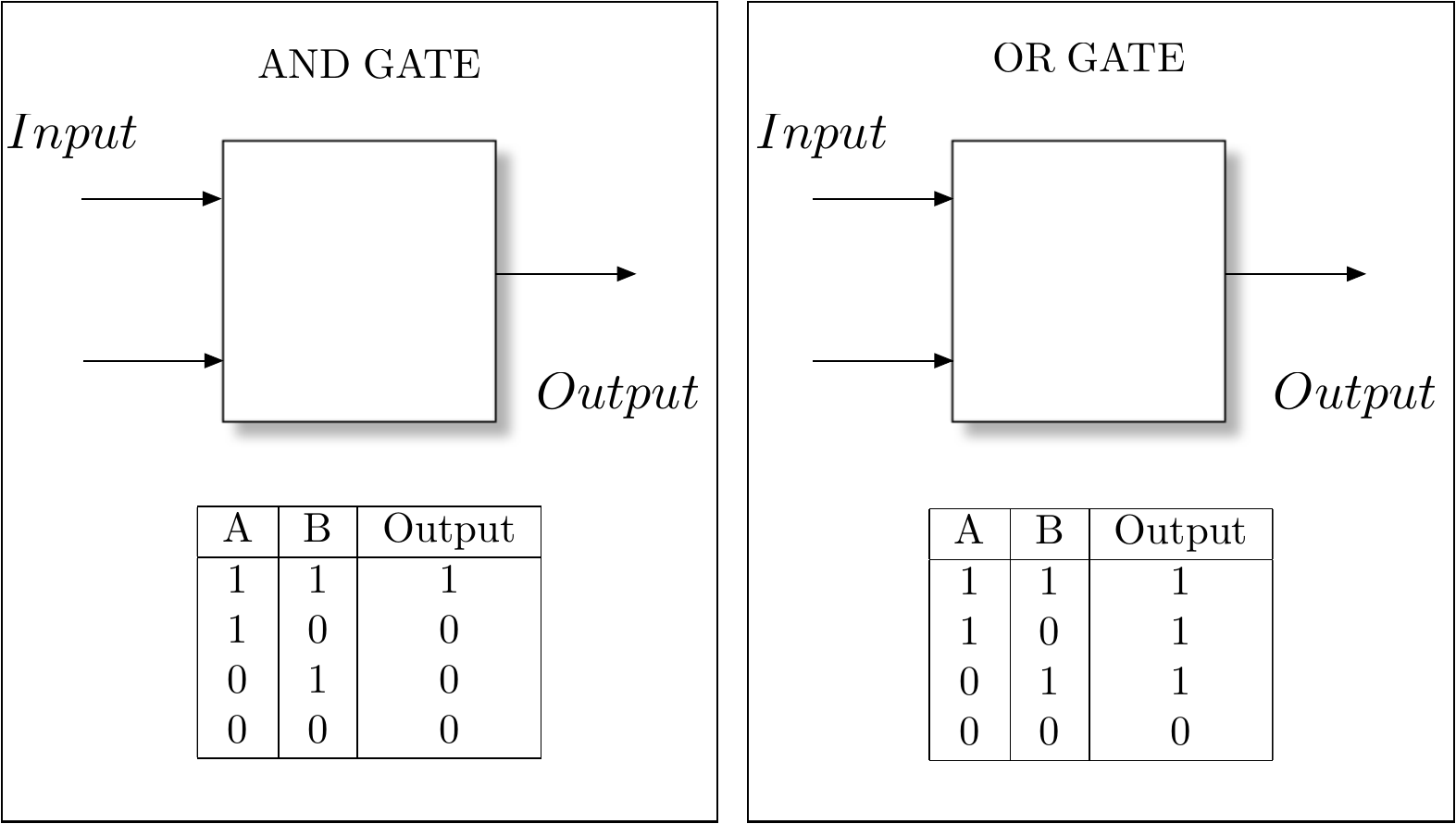}
		\caption{A computation is said to be logically irreversible if the input cannot be univocally defined only with the knowledge of the output. Here we have two examples of simple logic gates performing irreversible computations. The {\bf AND} gate (which corresponds to the logic connective $\wedge$) has the truth table shown on the bottom. Although the output $1$ can be uniquely obtained through the input (1, 1) as input, the input for the output 0 can be either (1, 0),(0, 1) or (0, 0). The presence of a 0 in the output is not enough to properly revert the computational process, and therefore we need an amount of extra information if we want to know the inputs with no errors. Therefore, the computations of this gate are {\em logically irreversible}. The same occurs with the {\bf OR} gate, corresponding to the logical connective $\vee$. This example has been taken from \cite{bennettlandauer1985}.}
	\label{fig:LogicGates}
\end{figure}

As remarked by \cite{bennettlandauer1985}, no computation ever generates information since the output is implicit in the input. However many operations destroy information whenever two previously distinct situations become indistinguishable, in which case the input contains more information than the output. For instance, if we consider the operation + defined on $\mathbb{N}$, the output 5 can be obtained from the following inputs: (0, 5), (1, 4), (2, 3), (3, 2), (4, 1), (5, 0). The concept of logical irreversibility is introduced in \cite[p. 264]{landauer1961} in order to study those computations for which its input cannot be unequivocally determined from its output:

\begin{quote}
``We shall call a device {\em logically irreversible} if the output of a
device does not uniquely define the inputs''.
\end{quote}
Conversely,  a device  is logically  reversible if  its output  can be
unequivocally defined from  the inputs -see figure (\ref{fig:LogicGates}). 

\subsection{\textit{Logical (ir)reversibility in terms of Turing machines}}
\label{TM}
In this subsection we define logically (ir)reversible computations in terms of a Turing machine. 

We informally describe, as usual, the action of a Turing machine on an infinite tape divided into squares as a sequence of simple operations that take place after an initial moment. At each step, the machine is at a particular internal state and its head examines a square of the tape (i.e., reads the symbol written on a square). The machine subsequently writes a symbol on that square, changes its internal state and moves the head one square leftwards or rightwards or remains at the same square.

Formally, a deterministic Turing machine $\mathcal{TM}$ is a triple composed of a finite set of internal states \textit{Q}, a finite set of symbols $\Sigma$ (an alphabet) and a transition function $\delta$,
\[
\mathcal{TM} = (Q, \Sigma, \delta).
\]
There is an initial state \textit{s} belonging to \textit{Q} and in general $\textit{Q} \cap \Sigma$ = $\emptyset$. Two special symbols, the blank $\sqcup$ and the initial symbol $\vartriangleright$, belong to $\Sigma$. Additionally
the transition function $\delta$ is understood as follows: 
\[
\delta: \textit{Q} \times \Sigma \rightarrow \textit{Q} \times \Sigma \times \left\{\textit{L}, \textit{R}, \square\right\},
\]
where `\textit{L}' and `\textit{R}' mean, respectively, ``move the head one square leftwards or rightwards'', and `$\square$' means ``stay at the square just examined''.

Thus, $\delta$ is the program of the machine; it specifies, for each combination of current state $\sigma_i \in \textit{Q}$ and current symbol $r_k \in \Sigma$, a triple
\[
\langle\sigma_j, r_{\ell}, \textit{D}\rangle,
\]
where $\sigma_j$ is the step immediately after $\sigma_i$, $r_{\ell}$ is the symbol to be overwritten on $r_k$, and $\textit{D}$ $\in$ $\left\{\textit{L}, \textit{R}, \square\right\}$.
\label{TM}
\begin{figure}
	\centering
		\includegraphics[scale=0.34]{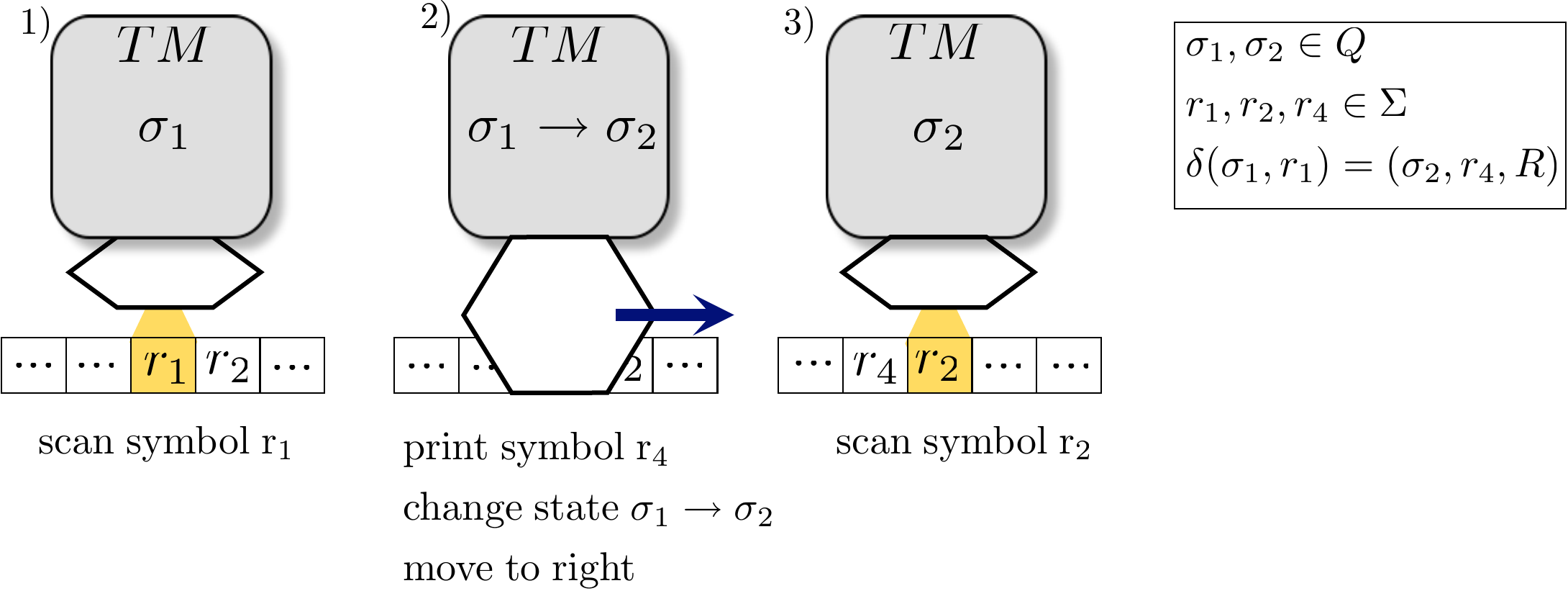}
		\caption{The sequence of a given computation in a Turing machine.}
	\label{fig:MT}
\end{figure}
A schema of how a computation is performed in a Turing machine is shown in figure (\ref{fig:MT}).\footnote{If $\delta$ is not a function from $\textit{Q} \times \Sigma$ to $\textit{Q}\times \Sigma \times \left\{\textit{L}, \textit{R}, \square\right\}$ but rather a subset of ($\textit{Q} \times \Sigma \times \textit{Q} \times \Sigma \times \left\{\textit{L}, \textit{R}, \square\right\})$, then $\mathcal{TM}$ is non-deterministic. This means that non-deterministic $\mathcal{TM}$s differ from deterministic $\mathcal{TM}$s in allowing for the possibility of assigning different outputs to one input. For simplicity we will consider in our argumentation only deterministic  $\mathcal{TM}$s. Note that this does not entail any loss of generality, since all non-deterministic  $\mathcal{TM}$s can be simulated by a deterministic $\mathcal{TM}$, although it seems that the deterministic $\mathcal{TM}$ requires exponentially many steps in \textit{n} to simulate a computation of \textit{n} steps by a non-deterministic $\mathcal{TM}$ (cfr. \cite[p. 221-227]{lewis97}).}

In general, we say that $\mathcal{TM}$ performs only {\em logically reversible computations} if the inverse function of $\delta$, $\delta^{-1}$, defined as
\[
\delta^{-1}: \textit{Q} \times \Sigma \times \left\{\textit{L}, \textit{R}, \square\right\} 
\rightarrow\textit{Q} \times \Sigma,\nonumber
\]
exists. This implies that for any input we have a different output and, therefore, we can invert the process for every element of the input set. If $\delta^{-1}$ does not exist then 
\[
(\exists \alpha, \beta\in \textit{Q} \times \Sigma):((\alpha\neq \beta)\wedge(\delta(\alpha)=\delta(\beta)\in  \textit{Q} \times \Sigma\times \left\{\textit{L}, \textit{R}, \square\right\})).
\]
Therefore, from the knowledge of $\gamma=\delta(\alpha)=\delta(\beta)$ we cannot determine with certainty whether the actual value of the input was either $\alpha$ or $\beta$. 

After these general definitions, we shall provide a particular definition of a Turing machine suitable for the study of the coding process. This coding machine $\mathcal{T}$ will be compounded of a set of internal states $\textit{Q}$, a transition function $\delta$ and two alphabets --an input alphabet  $\Omega=\{m_1,...,m_n\}$ and an output alphabet $\textit{S}=\{s_1,...,s_m\}$:   
\[
\mathcal{T} = (Q, \Omega, S, \delta).
\]
We shall call the elements of $\Omega$ referents and the elements of $\textit{S}$ signs. We assume that  $\textit{Q} \cap S$ = $\emptyset$ and that $\textit{Q} \cap \Omega$ = $\emptyset$. For simplicity, we also assume that, in  a coding process, the two alphabets are disjoint,
\[
\Omega \bigcap \textit{S} = \emptyset,
\]
 i.e., an object is either a sign or a referent, but not both. Technically, this implies that $\mathcal{T}$ can never reexamine a square where a sign has been printed, which means that $\mathcal{T}$ must move always in the same direction; assume for concreteness that it must move rightwards. Accordingly, we define the transition function as follows:

\begin{equation}
\delta: \textit{Q} \times \Omega \rightarrow \textit{Q} \times S \times \left\{\textit{R}\right\}
\label{eq:deltaDef}
\end{equation}
In order to clearly identify input and output configurations, we express the applications of $\delta$ in the following terms:
\[
\sigma_k m_i \rightarrow \sigma_js_{\ell}R,
\]
where $\sigma_k m_i$ and  $\sigma_js_{\ell}R$, ($\sigma_k, \sigma_j\in Q$, $m_i\in \Omega$, $s_{\ell}\in S$) are respectively input and output configurations.   

A coding machine $\mathcal{T}$ will perform solely logically reversible computations if there exists the inverse function of $\delta$, $\delta^{-1}$, defined as
\[
\delta^{-1}: \textit{Q} \times S \times \left\{\textit{L}\right\} 
\rightarrow\textit{Q} \times \Omega.\nonumber
\]
This implies that for any input configuration we have a different output configuration and, therefore, we can invert the process for every element of $\Omega$. The inexistence of $\delta^{-1}$ is due to the fact that
\[
(\exists \alpha, \beta\in \textit{Q} \times \Omega):((\alpha\neq \beta)\wedge(\delta(\alpha)=\delta(\beta)\in  \textit{Q} \times S \times \left\{R\right\})).
\]
If a coding machine is logically irreversible then it generates signs which are ambiguous in the sense that they encode more than one referent.

As shown by \cite{bennett1973}, a logically irreversible Turing machine can always be made logically reversible at every step. Thus, logical irreversibility is not an essential property of computations. Regarding the general questions (\ref{length}) and (\ref{size}) above formulated, it is particularly relevant for our concerns that a logically reversible machine need not be much more complicated than the irreversible machine it is associated with: computations on a reversible machine take about twice as many steps as on the irreversible machine they simulate and do not require a remarkably larger computing device.\footnote{The basic idea is that an irreversible computer can always be made reversible by having it save all the information it would otherwise lose on a separate extra tape that is initially blank. As Bennett shows, this can be attained ``without inordinate
increase in machine complexity, number of steps, unwanted output, or temporary storage capacity''. We refer the interested reader to \cite{bennett1973} for a detailed proof and illustration of this result.}

Therefore, the study of the complexity of the computations of the coding device alone does not seem to offer a \textit{necessity argument} for the emergence of ambiguity but only a relatively weak \textit{plausibility argument}, since reversible computations do not need to be significantly more complex than irreversible computations. However, as we shall see, it is possible to obtain a strong necessity argument for the emergence of ambiguity if instead of considering the complexity of the coding machine in isolation we study how a coding machine interacts with a decoding machine in an optimal way. 

\subsection{\textit{Logical reversibility and ambiguous codes}}
\label{ambcodes}

Before proceeding further in studying the concepts of logical (ir)reversibility in a communication system formed by two agents (a coder and a decoder) and the channel,  some clarifications are  in order. Firstly, we  note that
logical  reversibility refers  to  the  potential   existence  of  a
reconstruction or decoding algorithm, which does  not  entail  that,  in a  real
scenario,  such algorithm  is  at work;  in  other words,  logical
(ir)reversibility is  a feature  of the computations  alone. 
Secondly, in the process of transmitting a signal through a channel, the presence of noise in the channel through which the output is received may be responsible for the emergence of logical irreversibility. This implies that, although the coder agent could in principle compute in a reversible regime, the noise of the channel makes the cascade system \textit{coder agent + channel} analogue to a single computation device working in an irreversible regime. 
And finally, whereas logical (ir)reversibility is a property of the computational device (or coding algorithm) related to the potential existence of a reconstruction algorithm (or decoding algorithm), ambiguity is a property referred to signs: we say that a sign (an output of a coding computation) is ambiguous when the decoder can associate it with more than one referent (or input). A sign transmitted through a channel is ambiguous if the cascade \textit{coder agent + channel} is logically irreversible, which may be due to the computations of the coding agent itself or due to the noise of the channel.

\subsubsection{\textit{Noise: quantifying the degree of ambiguity}}
The minimal amount of additional information needed to properly reconstruct the input from the knowledge of the output is identified as the quantitative estimator of ambiguity. The more additional information we need, the more ambiguous the code is. This minimal amount of {\em dissipated} information is known as {\em noise} in standard information theory, and its formulation in terms of the problem we are dealing with is the objective of this subsection and the following one. 

To study logical irreversibility  in information-theoretical terms, we  choose the following simpler version of the transition function $\delta$ given in (\ref{eq:deltaDef}):  
\[
\delta:\Omega\to S.\nonumber
\]
This choice  puts aside the
role  of the  states but  is  justified for  the sake  of clarity  and
because the  qualitative nature  of the results  does not  change: the
only changes are the sizes of the input and output sets. Let $\delta_{ij}$ be a matrix by which
\[
\delta_{ij}=\left\{
\begin{array}{ll}
1\Leftrightarrow \delta(m_i)=s_j\\
0\;\mathrm{otherwise.}
\end{array}\nonumber
\right.
\]
Since the machine is deterministic, there is no possibility of having two outputs for a given input; therefore
\[
(\forall k\leq n)(\exists! i\leq m):\left[(\delta_{ki}=1)\wedge(\forall j\neq i)(\delta_{kj}=0)\right].
\]
In terms of probabilities, this deterministic behavior means that:
\begin{equation}
\mathbb{P}(s_k|m_i)=\delta_{ik}.
\label{eq:P(|)}
\end{equation}

To properly study the reversibility of the above defined coding machine $\mathcal{T}$, let us  define two  random variables, $X_{\Omega},  X_S$. $X_{\Omega}$
takes  values on the  set $\Omega$  following the  probability measure
$p$, being $p(m_k)$ the probability  to have symbol $m_k$ as the input
in a given  computation. Essentially, $X_{\Omega}$ describes the behavior of a fluctuating environment. $X_S$ takes values on  ${S}$ and follows
the probability  distribution $q$, which  for a given  $s_i\in {
  S}$, takes the following value:
\begin{equation}
q(s_i)=\sum_{k\leq n}p(m_k)\delta_{ki},
\label{eq:q}
\end{equation}
i.e.,  the probability of obtaining symbol  $s_i$  as the  output of  a
computation.  The amount of {\em uncertainty} in recovering the inputs
from the  knowledge of  the outputs of  the computations  performed by
$\mathcal{T}$ is related to logical irreversibility. In fact, this amount of {\em uncertainty} is precisely the amount of extra information we need to introduce to have a non-ambiguous code. This amount of
conditional uncertainty or extra information needed is well defined by the uncertainty function or
Shannon's conditional entropy:\footnote{Throughout the paper, $\log\equiv \log_2$.}
\begin{eqnarray}
H(X_{\Omega}|X_S)
=-\sum_{k\leq m} q(s_k)\sum_{i\leq n} \mathbb{P}(m_i|s_k)\log\mathbb{P}(m_i|s_k).
\label{eq:noise}
\end{eqnarray}
To properly derive $\mathbb{P}(m_i|s_k)$ in terms of the transition function of the coding machine we know, virtue of Bayes' theorem, that:
\[
\mathbb{P}(m_i|s_k)q(s_k)=\mathbb{P}(s_k|m_i)p(m_i),
\]
which, using equations (\ref{eq:P(|)}, \ref{eq:q}) leads us to a general expression only depending on the transition function of the coding machine and the prior probabilities $p$:
\[
\mathbb{P}(m_i|s_k)= p(m_i)\delta_{ik}\left(\sum_{j\leq n}\delta_{jk}p(m_j) \right)^{-1}.
\]
We observe that in the special case where $(\forall m_i)(p(m_i)=\frac{1}{n})$, the above equation simply reads:
\[
\mathbb{P}(m_i|s_k)= \left(\sum_{i\leq n}\delta_{jk} \right)^{-1}\delta_{jk}.
\]
Equation (\ref{eq:noise}) is the amount of {\em noise}, i.e., the information that is {\em dissipated} during the communicative exchange or, conversely, the (minimum) amount of information we need to externally provide for the system in order to perfectly reconstruct the input. Consistently with this interpretation, the amount of information shared by $X_{\Omega}$ and $X_{S}$, to be written as $I(X_S;X_{\Omega})$, will be:
\begin{equation}
I(X_S;X_{\Omega})=H(X_{\Omega})-H(X_{\Omega}|X_S),
\label{ShannonInfo}
\end{equation}
which is the so-called {\em Shannon Information} or  {\em Mutual Information} among the two random variables $X_S, X_{\Omega}$ \cite{Thomas:2001}, \cite{Ash:1990}.
In our particular case, such measure quantifies the amount of information we have from the input set by the only knowledge of the output set after the computations.

\subsubsection{\textit{Ambiguity and logical irreversibility}}

The interpretation  we provided for the noise equation enables us to connect ambiguity and logical (ir)reversibility. First, we emphasize a crucial fact: by the properties of Shannon's entropy,
\[
H(X_{\Omega}|X_S)\geq 0,
\]
which explicitly states  that information can be either {\em destroyed} or {\em maintained} but never {\em created} in the course of a given computation --as pointed out in \cite{bennettlandauer1985}.

If there is no uncertainty  in defining the input signals by the
only   knowledge    of   the   outputs, then
\[
H(X_{\Omega}|X_S)=0,
\]
i.e., there is certainty when reversing the computations performed by the coding machine. Therefore, the computations performed by $\mathcal{T}$ to define the code {\em are logically reversible} and the code {\em is not ambiguous}. 
Otherwise,   if   
\[
H(X_{\Omega}|X_S)>0,
\]
then, we need extra information (at least $H(X_{\Omega}|X_S)$) to properly reverse the process, which indicates that the computations defining the code are {\em logically irreversible} and, thus, that {\em the code is ambiguous}. 

We have therefore identified in a quantitative way the ambiguity of the code with {\em the amount of uncertainty of the reversal of the coding process} or the minimal amount of {\em additional information we need to properly reverse the coding process}. Furthermore, we have identified the source of uncertainty through the concept of {\em logical irreversibility}, which is a feature of the computations generating the code. In this way, we establish the following correspondences:
\[
\mathrm{logically\;reversible\;computations\;\Leftrightarrow \;No\;Ambiguity}\;\Leftrightarrow H(X_{\Omega}|X_S)=0
\]
\[
\mathrm{logically\;irreversible\;computations\;\Leftrightarrow \;Ambiguity}\;\Leftrightarrow\;H(X_{\Omega}|X_S)>0
\]
\[
\mathrm{Amount\;of\;Ambiguity\;}=H(X_{\Omega}|X_S).
\]

Now that we have rigorously defined ambiguity on theoretical grounds of computation theory and information theory, we are ready to explain why it appears in natural communication systems. As we shall see in the following section, the reason is that natural systems must satisfy certain constraints that generate a communicative tension whose solution implies the emergence of a certain amount of ambiguity.

\section{The emergence of ambiguity in natural communication}
\label{LEP}

The tension we referred to at the end of the last section was postulated by the linguist G. K. Zipf \cite{zipf1965} as the origin of the widespread scaling behavior of word appearance having his name. Such a communicative tension was conceived in terms of a balance between two opposite {\em forces}: \textit{the speaker's economy force} and \textit{the auditor's economy force}.

\subsection{\textit{Zipf's hypothesis}}
Let us thus informally present Zipf's vocabulary balance between two opposite forces, the speaker's economy force and the auditor's economy force \cite[pp. 19-31]{zipf1965}. The speaker's economy force (also called unification force) is conceived as a tendency ``to reduce the size of the vocabulary to a single word by unifying all meanings'', whereas the auditor's economy force (or diversification force) ``will tend to increase the size of a vocabulary to a point where there will be a distinctly different word for each different meaning''. Therefore, a {\em conflict} will 
be present while trying to simultaneously minimize these two theoretical opposite forces, and the resulting vocabulary will emerge from a \textit{cooperative solution} to that conflict. In Zipf's words, 
\begin{quote}
``whenever a person uses words to convey meanings he will automatically try to get his ideas across most efficiently by seeking a balance between the economy of a small wieldy vocabulary of more general reference on the one hand, and the economy of a larger one of more precise reference on the other, with the result that the vocabulary of \textit{n} different words in his resulting flow of speech will represent a \emph{vocabulary balance} between our theoretical Forces of Unification and Diversification'' \cite[p. 22]{zipf1965}.
\end{quote}
Obviously the unification force ensures a minimal amount of \emph{lexical} ambiguity, since it will require some words to convey more than one meaning, and the diversification force constrains such amount. Thus, lexical ambiguity can be viewed as a consequence of the vocabulary balance. Although Zipf's vocabulary balance, as stated, provides a useful intuition to understand the emergence of lexical ambiguity by emphasizing the cooperative strategy between communicative agents, it lacks the necessary generality to provide a principled account for the origins of ambiguity beyond the particular case of lexical ambiguity. In the following sections we shall present several well-known concepts in order to generalize Zipf's informal condition and provide solid foundations for it.

We remark that Zipf conceived the vocabulary balance as a particular case of a more general principle, the Least Effort Principle, ``the primary principle that governs our entire individual and collective behaviour of all sorts, including the behaviour of our language and preconceptions''. In Zipf's terms, 
\begin{quote}
``the Principle of Least Effort means, for example, that a person in solving his immediate problems will view these against  the background of his probable future problems as \emph{estimated by himself}. Moreover he will strive to solve his problems in such a way as to minimize the total work that must be expended in solving both his immediate problems \emph{and} his probable future problems. That in turn means that the person will strive to minimize the \emph{probable average rate of his work-expenditure} (over time). And in so doing he will be minimizing his \textit{effort}, by our definition of effort. Least effort, therefore, is a variant of least work'' \cite[p. 22]{zipf1965}.
\end{quote}
Hence, we consider the symmetry equation between the complexities of the coder and the decoder we shall arrive at to be a particular instance of the Least Effort Principle. 

\subsection{\textit{Symmetry in coding/decoding complexities}}
\label{Kolmogorov}

How can we accommodate the previous intuitions to the formal framework proposed in section \ref{revers}? 
The auditor's economy force leads to a one-to-one mapping between
${\Omega}$ and ${\cal S}$. In this case, the computations performed by $\mathcal{T}$ to generate the code are logically reversible and thus generate an unambiguous
code,  and  no supplementary  amount  of  information to  successfully
reconstruct ${X_\Omega}$ is required. However, the speaker's economy force conspires exactly in the opposite direction. In these latter terms, the best option is an {\em all-to-one} mapping, i.e., a coding process where any realization of $X_{\Omega}$ is coded through a single signal. The coding computations performed by $\mathcal{T}$  are logically irreversible and the generated code is ambiguous, for it is clear that the knowledge of the output tells us nothing about the input. In order to characterize this 
conflict, let us properly
formalize the  above intuitive statement:  the auditor's force  pushes the code in such a way that it is possible to
reconstruct  $X_{\Omega}$  through the  intermediation  of the  coding
performed by $\mathcal{T}$.  Therefore, the  amount of {\em bits} the
decoder of $X_S$ needs to unambiguously reconstruct $X_{\Omega}$ is
\[
H(X_{\Omega}, X_S)=-\sum_{i\leq n} \sum_{k\leq n}\mathbb{P}(m_i, s_k)\log \mathbb{P}(m_i, s_k),\nonumber
\]
which is the {\em joint Shannon entropy} or, simply, {\em joint entropy} of the two random variables $X_{\Omega}, X_S$ \cite{Thomas:2001}. 
From the codification process, the auditor receives $H(X_S)$ bits, and
thus, the remaining uncertainty it must face will be
\[
H(X_{\Omega}, X_S)-H(X_S)=H(X_{\Omega}|X_S),
\]
where
\[
H(X_S)=-\sum_{i\leq n}q(s_i)\log q(s_i),
\]
(i.e, the {\em entropy} of the random variable $X_S$) and
\[
H(X_{\Omega}|X_S)=-\sum_{i\leq n}q(s_i)\sum_{k\leq n}\mathbb{P}(m_k|s_i)\log \mathbb{P}(m_k|s_i),
\]
the {\em conditional entropy} of the random variable $X_{\Omega}$ conditioned to the random variable $X_S$.
At this point Zipf's hypothesis becomes crucial. Under this interpretation, the tension between the auditor's force and the speaker's force is cooperatively solved by imposing a 
 symmetric  balance  between the  {\em efforts} associated to each communicative agent:  the
 coder sends as many bits as  the additional bits the decoder needs to
 perfectly reconstruct $X_{\Omega}$:
\begin{equation}
H(X_S)=H(X_{\Omega}|X_S).
\label{Symmetryeq}
\end{equation}
This is the {\em symmetry equation} governing the communication among cooperative agents when we take into account computational efforts --which have been associated here with the entropy or {\em complexity} of the code.\footnote{In the context of this section, {\em complexity} has to be understood in the sense of {\em Kolmogorov complexity}. Given an abstract object, such a general complexity measure is the {\em length, in bits, of the minimal program whose execution in a Universal Turing machine generates a complete description of the object}. In the case of codes where the presence of a given signal is governed by a probabilistic process, it can be shown that Kolmogorov complexity equals (up to an additive constant factor) the entropy of the code \cite{Thomas:2001}.} Selective pressures will push $H(X_S)$ and, at the same time, by equation (\ref{Symmetryeq}), the amount of ambiguity will also grow, as a consequence of the cooperative nature of communication.\footnote{Equations of this kind have been obtained in the past through different approraches; cfr. \cite{Harremoes:2001} and \cite{Ferrer:2003}.} 

Equation (\ref{Symmetryeq}) specifies that a certain amount of information must be lost (or equivalently, a certain amount of ambiguity must appear) if coder and decoder miminize their effors in a symmetric scenario. A further question is how much information is lost due to equation (\ref{Symmetryeq}).
In order to measure this amount of information, we must take into consideration the properties of the  {\em Shannon Information} or  {\em Mutual Information} among the two random variables $X_S, X_{\Omega}$, defined in equation (\ref{ShannonInfo}).
An interesting property of Shannon information is its symmetrical behavior, i.e., 
$I(X_S;X_{\Omega})=I(X_{\Omega}; X_S)$ \cite{Thomas:2001}, \cite{Ash:1990}. Thus, by equation (\ref{ShannonInfo}), 
\[
H(X_{\Omega})-H(X_{\Omega}|X_S)=H(X_S)-H(X_S|X_{\Omega}),
\]
where $H(X_s|X_{\Omega})=0$, because the Turing machine is deterministic.\footnote{Notice that, if the Turing machine is deterministic, every input generates one and only one output. The problem may arise during the reversion process, if the computations are logically irreversible.} Therefore, by applying directly equation (\ref{Symmetryeq}) to the above equation we reach the following identity:
\begin{equation}
H(X_S)=\frac{1}{2}H(X_{\Omega}).
\label{eq:1/2}
\end{equation}
Thus,
\begin{eqnarray}
I(X_S;X_{\Omega})&=&H(X_{\Omega})-H(X_{\Omega}|X_S)\nonumber\\
^{by\;eq.\;(\ref{Symmetryeq})}&=&H(X_{\Omega})-H(X_S)\nonumber\\
^{by\;eq.\;(\ref{eq:1/2})}&=&\frac{1}{2}H(X_{\Omega}).\label{der}
\end{eqnarray}
The above derivation shows that half of the information is dissipated during the communicative exchange if coding and decoding computations are symmetrically or cooperatively optimized. Accordingly, an amount of ambiguity must emerge. Ambiguity is not an inherent imperfection of a communication system or a footprint of poor design, but rather a property emerging from conditions on efficient computation: coding and decoding computations have a cost when they are performed by physical agents and thereby it becomes crucial to minimize the costs of coding and decoding processes. Whereas studying the process of an isolated coding agent would not provide a necessity argument for the emergence of ambiguous codes (as noted in section \ref{TM}, following \cite{bennett1973}), a formalization of an appropriately general version of Zipf's intuitions along the course we have developed provides a general necessity argument for the emergence of ambiguity. 

Note that he derivation in (\ref{der}) has been performed by assuming that there is no noise affecting the process of output set observation. If we assume the more realistic situation in which there is noise in the process of output observation, the situation is even worse, and, actually, $I(X_s;X_{\Omega})=\frac{1}{2}H(X_{\Omega})$ would be considered as an upper bound; therefore, in presence of noise in the process of output observation, this equation must be replaced by:
\[
I(X_S;X_{\Omega})<\frac{1}{2}H(X_{\Omega}).
\]

We will end this section by finding a bound on the {\em wrong messages} one can expect for an ambiguous code arising from equation (\ref{Symmetryeq}). Let $p_e$ be the probability of wrongly decoding a given signal --which, as shown by the fact that $H(X_{\Omega}|X_S)>0$, will necessary happen. Let $h(p_e)$ be the following binary entropy:
\[
h(e)=-p_e\log p_e-(1-p_e)\log(1-p_e),
\]
then, by the so-called {\em Fano's inequality}, we have that:
\[
p_e\geq \frac{H(X_{\Omega}|X_S)-h(e)}{\log (n-1)}.
\]
Therefore,
\begin{eqnarray}
p_e&\geq& \frac{H(X_{\Omega}|X_S)-h(e)}{\log (n-1)}\nonumber\\
^{by\;eq.\;(\ref{Symmetryeq})}&=&\frac{H(X_S)-h(e)}{\log (n-1)}\nonumber\\
^{by\;eq.\;(\ref{eq:1/2})}&=&\frac{H(X_{\Omega})-2h(e)}{2\log (n-1)}.\nonumber
\end{eqnarray}
The above equation can be rewritten, in the paradigmatic case in which
\[
(\forall m_i)\left(p(m_i)=\frac{1}{n}\right),
\]
as
\[
p_e\geq \frac{1}{2},
\]
since $H(_{\Omega})=\log n$. This means that, in this case, {\em at least} one message out of two will be wrongly decoded or, in other words, in the kind of ambiguous codes we described and under the condition of equiprobability among the elements of $\Omega$, the correct decoding of more than half of the sent messages will depend on some external, additional source of information.

\section{Discussion}
\label{Discussion}

In this article we have constructed a communicative argument based on fundamental concepts from computation theory and information theory in order to understand the emergence of ambiguity.

We have identified the source of ambiguity in a code with the concept of logical irreversibility in such a way that a code is ambiguous when the coding process performs logically irreversible computations. Since logical irreversibility is not an essential property of computations and a logically reversible machine need not be much more complicated than the logically irreversible machine it simulates, we have inquired into how a coding machine interacts with a decoding machine in an optimal way in order to identify the source of ambiguity. We have quantified the ambiguity of a code in terms of the amount of uncertainty of the reversal of the coding process, and we have subsequently formulated the intuition that coder and decoder cooperate in order to minimize their efforts in terms of a symmetry equation that forces the coder to send only as many bits as the additional bits the decoder needs to perfectly reconstruct the coding process. Given the symmetric behaviour of Shannon information it has been possible to quantify the amount of ambiguity that must emerge from the symmetry equation regardless the presence of noise in the channel: at least half of the information is dissipated during the communicative process if both the coding and the decoding computations are cooperatively minimized. As noted explicitly in Appendix \ref{Inequalities}, the presence of ambiguity associated to a computational process realized by a physical system seems as necessary as the generation of heat during a thermodynamical process. 

The interest of the symmetry equation from which we derive a certain amount of ambiguity in natural languages is further corroborated in \cite{corominas2011}. In this study it is shown that Zipf's law emerges from two factors: a static symmetry equation that solves the tension between coder and decoder (namely, our symmetry equation \ref{Symmetryeq}) and the path-dependence of the code evolution through time, which is mathematically stated by imposing a variational principle between successive states of the code (namely, Kullback's Minimum Discrimination of Information Principle). We thus  conclude this study by emphasizing the importance of the symmetry equation for the understanding of how communicative efficiency considerations shape linguistic productions.

\appendix

\section{Appendix: Ambiguity and physical irreversibility}
\label{Inequalities}

Throughout the paper we have highlighted the strict relation between the logical irreversibility of the computations generating a given code and the ambiguity of the latter. In this appendix we briefly revise the role of logical irreversibility in the foundations of physics, through its relation to thermodynamic irreversibility. 

Our objective here is to show, in a rather informal way, that reasonings and concepts that are commonly used in physics can be naturally connected to the conceptual development of this paper. We warn the reader that this appendix does not attempt to provide formal relations between ambiguous codes and its energetic cost or other thermodynamic quantities, nor to relate the emergence of ambiguity to some explicit physical process. The rigorous exploration of the above mentioned topics is a fascinating issue, but lies beyond the scope of this paper.

The strict relation of thermodynamic irreversibility and logical irreversibility is a hot topic of debate since the definition of the equivalence of heat and bits by R. Landauer \cite{landauer1961}. This equivalence, known as {\em Landauer's principle}, states that, for any {\em erased} bit of information, a quantity of at least
\[
k_BT\ln 2
\]
joules is dissipated in terms of heat, being $k_B=1.38\times 10^{-23}J/K$ the Boltzmann constant and $T$ the temperature of the system. This principle relates logical irreversibility and thermodynamical irreversibility; however, it is worth noting that it provides only a lower bound and that it is far away from the energetic costs of any real computing process. 
To see how we can connect both types of irreversibility, we first state that thermodynamical irreversibility is a property of abstract processes --interestingly, almost all processes taking place in our everyday life are irreversible. 
The common property of such processes is that they {\em generate} thermodynamical entropy. The second law of thermodynamics states that any physical process generates a non-negative amount of entropy; i.e., for the process $\mathbf{P}$,
\[
\Delta S(\mathbf{P})\geq 0.
\]
The units of physical entropy are {\em nats} instead of bits. Now suppose that we face the problem of reversing the process $\mathbf{P}$ --for example, a gas expansion-- by which $\Delta S(\mathbf{P})> 0$. Without further help, the reversion of this process is {\em forbidden} by the second law, since it would generate a net amount of {\em negative} entropy. Therefore, we will need external energy to reverse the process. Similarly, we have observed that
\[
H(X_{\Omega}|X_s)\geq 0,
\]
which means that information cannot be created during an information process. A negative amount of $H(X_{\Omega}|X_s)$ would imply, by virtue of equation (\ref{ShannonInfo}), a net creation of information. Therefore, we face the same problem. Indeed, if we have a computational process $\mathbf{C}$ by which $H_{\mathbf{C}}(X_{\Omega}|X_s)>0$, the reversion of such a process, with no further external help, would be a process by which the computations would generate information. The reversion, as we have discussed above, is only possible by the external addition of information. 
Thus the information {\em flux} can only be maintained (in the case where all computations are logically reversible) or degraded, and the same applies for the energy flux: by the second law, the energy flux can only be maintained (in the case of thermodynamically reversible processes) or degraded. 

We can informally find a quantitative connection between the two entropies. If $\mathbf{Q}(\mathbf{P})$ is the heat generated during the physical process, its associated physical entropy generation is defined as 
\[
\Delta S(\mathbf{P})=\frac{\mathbf{Q}(\mathbf{P})}{T}.
\]
In turn, if we consider an ideal --from the energetic viewpoint-- computational process $\mathbf{C}$, we know, from Landauer's principle, that 
\[
\mathbf{Q}(\mathbf{C})=k_BT\ln 2\times {\rm erased\;bits}.
\]
And we actually know how many bits have been erased --or dissipated. Exactly $H(X_{\Omega}|X_s)$ bits. Therefore, the physical entropy generated by this ideal, irreversible computing process will be:
\[
\Delta S(\mathbf{C})= k_B\ln 2 H_{\mathbf C}(X_{\Omega}|X_s).
\]
Accordingly, {\em logically irreversible computations are thermodynamically irreversible}. Notice that this only proves the implication between both irreversibilities, but is of no practical use, since it is a lower bound. Any real computing process will be such that:
\[
\Delta S(\mathbf{C})\gg k_B\ln 2 H_{\mathbf C}(X_{\Omega}|X_s).
\]
We finally highlight that one could object that, under the above considerations, the most favorable situation would be the one in which all computations are performed in a logically reversible way, since there would not be an energy penalty. However, this interpretation is misleading. Imagine a coding machine receiving an informational input but working in a totally irreversible way. In this case, there would be a dissipation of information that would undergo into heat production. The energy dissipated in terms of heat, however, would come from the environment, not from the machine, which would be only heated. The creation of a code in this machine to let information be coded and flow would imply, on the contrary, to {\em write} a code into the machine, and thus, to erase the initial configuration of the machine --whatever it was, maybe a random one--, to properly adapt it to a consistent coding process. We observe that this process would demand energy that would not come, at least directly, from the environment. Therefore, a perfect coding performing logically reversible computations only would be, in principle, energetically more demanding than a logically irreversible coding.
 
We insist that the above considerations fall into the abstract level and practical implementations must face multiple additional problems which have been not been taken into consideration. With this short exposition we only want to emphasize the general character of logical irreversibility and ambiguity in natural communication systems. More than an imperfection, ambiguity seems to be, for natural communication systems, a feature as unavoidable as the generation of heat during a thermodynamical process. 


%
%

\section*{Acknowledgements}
This work has been supported by the Secretary for Universities and Research of the Ministry of Economy and Knowledge of the Government of Catalonia and the Cofund programme of the Marie Curie Actions of the 7th R\&D Framework Programme of the European Union, the research projects 2009SGR1079, FFI201123356 (JF) and the James S. McDonnell Foundation (BCM). The final version is published in \textit{The Journal of Logic, Language and Information} and is available at http://link.springer.com/. We would like to thank the members of the Centre de Ling\"u\'istica Te\`orica that attended the course on ambiguity for postgraduate students we taught within the PhD program on cognitive scicence and language (fall semester, 2010). We are especially grateful to M. Teresa Espinal for many interesting discussions during the elaboration process of this study and to Adriana Fasanella, Carlos Rubio, Francesc-Josep Torres and Ricard Sol\'e for carefully reading a first version of this article and providing us with multiple improvements. We also wish to express our gratitude to two anonymous reviewers for several remarks that helped us to clarify and strengthen our proposal.

\bibliographystyle{spmpsci}      


\end{document}